\documentclass[11pt]{article}
\usepackage[usenames,dvipsnames]{color}
\usepackage{times}
\usepackage{hyperref}
\usepackage{url}
\usepackage{latexsym}
\usepackage{amsthm}
\usepackage{multirow}
\usepackage{graphicx}
\usepackage{amsfonts,eucal,amsbsy}
\usepackage{amsmath}
\usepackage{amssymb}
\usepackage{sidecap}
\usepackage{microtype}
\usepackage{mdwlist}
\usepackage{algorithm}
\usepackage{algorithmic}
\usepackage{subfigure}
\usepackage{mathtools}
\usepackage{acl2015}

\DeclareMathOperator*{\argmax}{arg\,max}

\newcommand{\vect}[1]{\mathbf{#1}}

\usepackage{enumitem}

\setitemize{noitemsep,topsep=6pt,parsep=0pt,partopsep=0pt,itemindent=0pt,leftmargin=12pt,labelwidth=12pt}
\setenumerate{noitemsep,topsep=6pt,parsep=0pt,partopsep=0pt,itemindent=0pt,leftmargin=12pt,labelwidth=12pt}

\usepackage[size=small,labelfont=bf]{caption}

\title{Bayesian Optimization of Text Representations}

\author{Dani Yogatama \\
  Language Technologies Institute \\
  School of Computer Science \\
  Carnegie Mellon University \\
 Pittsburgh, PA 15213, USA \\
  {\tt dyogatama@cs.cmu.edu} \\\And
  Noah A. Smith \\
  Language Technologies Institute \\
  School of Computer Science \\
  Carnegie Mellon University \\
 Pittsburgh, PA 15213, USA \\
  {\tt nasmith@cs.cmu.edu} \\}

\date{}

\begin{document}
\maketitle
\begin{abstract}
When applying machine learning to problems in NLP, there are many
choices to make about how to represent input texts.  These choices can
have a big effect on performance, but they are often uninteresting
to researchers or practitioners who simply need a module that performs
well.  We propose an approach to optimizing over this space of
choices,
formulating the problem as global optimization.
  We apply a sequential model-based optimization technique and 
 show that our method makes standard linear models  
  competitive with 
  more sophisticated, expensive state-of-the-art methods based on latent variable models
  or neural networks on various topic classification and sentiment analysis problems.
Our approach is a first step towards black-box NLP systems that
work with raw text and do not require manual tuning.
\end{abstract}

\section{Introduction}
NLP researchers and practitioners spend a considerable amount of time
comparing machine-learned models of text that differ in relatively
uninteresting ways.  For example, in categorizing texts, should the
``bag of words'' include bigrams, and is tf-idf weighting a good idea?
These choices matter experimentally, often leading to big differences
in performance, with little consistency across tasks and datasets in which
combination of choices works best.
Unfortunately, these differences
tell us little about language or the problems
that machine learners are supposed to solve.  

We propose that these decisions can be automated in a similar way to
hyperparameter selection (e.g., choosing the strength of a ridge or
lasso regularizer).  Given a particular
text dataset and classification task, we introduce a technique for
optimizing over the space of representational choices, along with
other ``nuisances'' that interact with these decisions, like hyperparameter selection.\footnote{In
  \S\ref{sec:discuss} we argue that the technique is also applicable
  in unsupervised settings.}  For example, using higher-order
$n$-grams means more features and a need for stronger regularization
and more training iterations.  Generally, these decisions about
instance representation are made by
humans, heuristically; our work is the first to automate them.

Our technique instantiates sequential model-based optimization (SMBO;
Hutter et al., 2011). \nocite{smbo}  SMBO and other Bayesian
optimization approaches have been shown
to work well for hyperparameter tuning
\cite{bergstra,portfolio,bayesiangp}.  Though popular in
computer vision \cite{makings}, these techniques have received little
attention in NLP. 

We apply the technique to
logistic regression on a range of topic and sentiment classification
tasks.   Consistently, our method finds representational
choices that perform better than linear baselines previously reported
in the literature, and that, in some cases, are competitive with more
sophisticated non-linear models trained using neural networks.

\section{Problem Formulation and Notation}

  Let the training data consist of  a collection of 
pairs $\boldsymbol{d}_{\mathit{train}} = \langle \langle d.i_1,
d.o_1\rangle, \ldots, \langle d.i_n, d.o_n\rangle\rangle$, where each
input $d.i \in \mathcal{I}$ is a
text document and each output $d.o \in \mathcal{O}$, the output space.
The overall training
goal is to maximize a performance function $f$ (e.g., classification
accuracy, log-likelihood, $F_1$ score, etc.) of a machine-learned
model, on a held-out dataset, $\boldsymbol{d}_{\mathit{dev}}  \in
(\mathcal{I} \times \mathcal{O})^{n'}$.  

Classfication proceeds in three steps:  first, $\vect{x}: \mathcal{I} \rightarrow
\mathbb{R}^N$ maps each input to a vector representation.
Second, a classifier is learned from the inputs (now transformed into
vectors) and outputs:  $L : (\mathbb{R}^N \times \mathcal{O})^n
\rightarrow (\mathbb{R}^N \rightarrow \mathcal{O})$.  Finally, the resulting
classifier $c: \mathcal{I} \rightarrow \mathcal{O}$ is fixed as 
\begin{align*}
\begin{array}{ccc} 
L(\boldsymbol{d}_{\mathit{train}}) &\circ&
\vect{x} \\
{\small \mathcal{O} \leftarrow \mathbb{R}^N} && {\small \mathbb{R}^N \leftarrow
\mathcal{I}}
\end{array}
\end{align*}
(i.e., the composition of the representation function with
the learned classifier).

Here we consider linear classifiers of the form
\begin{align}
c(d.i) &= \argmax_{o \in \mathcal{O}} \vect{w}_o^\top\vect{x}(d.i)
\end{align}
where the coefficients $ \vect{w}_o \in \mathbb{R}^N$, for each output
$o$, are learned using logistic
regression on the training data. We let $\vect{w}$ denote the
concatenation of all $\vect{w}_o$.   Hence the parameters can be
understood as a function of the training data and the representation
function $\vect{x}$.  The performance function $f$, in turn, is a
function of the held-out data $\boldsymbol{d}_{\mathit{dev}}$  and $\vect{x}$---also $\vect{w}$ and
$\boldsymbol{d}_{\mathit{train}}$, through $\vect{x}$. 
For simplicity, we will write ``$f(\vect{x})$'' when the rest are
clear from context.

Typically, $\vect{x}$ is fixed by the model designer, perhaps after
some experimentation, and learning focuses on selecting the parameters
$\vect{w}$.
For logistic regression and many other linear models, this training
step reduces to
convex optimization in $N |\mathcal{O}|$ dimensions---a solvable problem that is
still costly for  large datasets and/or large output spaces.   In seeking to maximize $f$ with
respect to $\vect{x}$, we do not wish to
carry out training any more times than necessary.

Choosing $\vect{x}$ can be understood as a problem of selecting
\emph{hyperparameter values}.  We therefore turn to 
Bayesian optimization, a family of techniques recently introduced for selecting
hyperparameter values intelligently when solving for parameters
($\vect{w}$) is costly.

\section{Bayesian Optimization}
Our approach is based on sequential model-based optimization (SMBO;
Hutter et al., 2011).  \nocite{smbo}
It iteratively chooses representation functions $\vect{x}$.  On each
round, it makes this choice through a nonparametrically-estimated
probabilistic model of $f$, then evaluates $f$---we call this a ``trial.''  
As in any iterative search algorithm, the goal is to balance exploration of
options for $\vect{x}$ with exploitation of previously-explored
options, so that a good choice is found in a small number of trials.
See Algorithm~\ref{alg:smbo}.

More concretely, in the $t$th trial, $\vect{x}_t$ is selected using an
acquisition function $\mathcal{A}$ and a ``surrogate'' probabilistic
model $p_t$.
Second, $f$ is evaluated given $\vect{x}_t$---an expensive operation which
involves training to select parameters $\vect{w}$ and assessing performance
on the held-out data.
Third,  the
probabilistic model is updated using a nonparametric estimator.

\begin{algorithm}[h]
   \caption{SMBO algorithm  
   \label{alg:smbo}}
\begin{algorithmic}
   \STATE {\bfseries Input:} number of trials $T$, target function $f$
   \STATE $p_{1} = \text{initial surrogate model}$
   \STATE Initialize $y^{\ast}$
   \FOR{$t=1$ {\bfseries to} $T$}
   \STATE $\vect{x}_t \leftarrow \argmax_{\vect{x}}$
   $\mathcal{A}(\vect{x}; p_t, y^\ast)$
   \STATE $y_t \leftarrow$ evaluate $f(\vect{x}_t)$
   \STATE Update $y^{\ast}$
   \STATE Estimate $p_t$ given $\vect{x}_{1:t}$ and $y_ {1:t}$
   \ENDFOR
\end{algorithmic}
\end{algorithm}

We next describe the acquisition function $\mathcal{A}$ and the
surrogate model $p_t$ used in our experiments.  

\subsection{Acquisition Function}

A good acquisition function returns 
high values for $\vect{x}$ such that either the value $f(\vect{x})$ is
predicted to be high, or because uncertainty about $f(\vect{x})$'s
value is high; balancing between these is the classic tradeoff between
exploitation and exploration.
We use a criterion called Expected Improvement (EI; Jones, 2001),
 \nocite{ei}
which is the expectation (under the current surrogate model
$p_t$) that the choice $y$ will exceed $y^\ast$:
\begin{align*}
\mathcal{A}(\vect{x}; p_t, y^{*}) = \int^{\infty}_{-\infty}
\max(y - y^{*}, 0) p_t(y \mid \vect{x}) dy
\end{align*}
where $y^\ast$ is chosen depending on the surrogate model, discussed
below.  (For now, think of it as a strongly-performing ``benchmark''
value of $f$, discovered in earlier iterations.)
Other options for the acquisition function include
maximum probability of improvement \cite{ei},
minimum conditional entropy \cite{villemonteix},
Gaussian process upper confidence bound \cite{ucb}, or a combination of them \cite{portfolio}.
We selected EI because it is the most widely used
acquisition function that has been shown to work
well on a range of tasks. 
 
\subsection{Surrogate Model}
\label{sec:surrogate}
As a surrogate model, we use a tree-structured Parzen estimator (TPE;
Bergstra et al., 2011). \nocite{bergstra}
This is a nonparametric approach to density estimation.  We seek
to estimate  $p_t(y \mid \vect{x})$ where $y =
f(\vect{x})$, the performance function that is expensive to compute
exactly.
The TPE approach is as follows:
\begin{align*}
p_t(y \mid \vect{x}) &\propto p_t(y) \cdot
p_t(\vect{x} \mid y) \\
p_t(\vect{x} \mid y) & =
\begin{cases}
p_t^{<}(\vect{x}),& \text{if } y < y^*\\
p_t^{\geq}(\vect{x}),& \text{if } y \geq y^*\\ 
\end{cases} 
\end{align*}
where $p_{t}^{<}$ and $p_t^{\geq}$ are densities estimated using
observations from previous trials that are less than and greater
than $y^*$, respectively.
In TPE, $y^\ast$ is defined as some quantile of the observed $y$; we use
15-quantiles.

As shown by
\newcite{bergstra}, the Expected Improvement in TPE can be written as:
\begin{align}
\label{eq:eitpe}
\mathcal{A}(\vect{x}; p_t, y^{*}) \propto \left( \gamma + \frac{p_t^{<}(\vect{x})}{p_t^{\geq}(\vect{x})}(1-\gamma) \right)^{-1},
\end{align}
where $\gamma = p_t(y < y^*)$, fixed at $0.15$ by definition of
$y^\ast$ (above). 
Here, we prefer $\vect{x}$ with high probability under
$p_t^{\geq}(\vect{x})$ and low probability under $p_t^{<}(\vect{x})$. 
To maximize this quantity, we draw many candidates
according to $p_t^{\geq}(\vect{x})$ and evaluate them
according to $p_t^{<}(\vect{x})/p_t^{\geq}(\vect{x}$).
Note that $p(y)$ does not need to be given an explicit
form.

In order to evaluate Eq.~\ref{eq:eitpe},
we need to compute 
$p_t^{<}(\vect{x})$ and $p_t^{\geq}(\vect{x})$.
These joint distributions depend on the graphical model
of the hyperparameter space---which is allowed to form a tree structure.

We discuss how to compute $p_t^{<}(\vect{x})$ in the following.
$p_t^{\geq}(\vect{x})$ is computed similarly, using trials where $y \geq y^{\ast}$.
We associate each hyperparameter with a node in the graphical model;
consider the $k$th dimension of $\vect{x}$, denoted by random variable $X^k$.
\begin{itemize}
\item If $X^k$ ranges over a discrete set $\mathcal{X}$, TPE uses a reweighted categorical
distribution, where the probability that $X^k = x$ is proportional
to a smoothing parameter plus the counts of
occurrences of $X^k = x$ in $\vect{x}^k_{1:t}$ 
with $y_t < y^{\ast}$.
\item When $X^k$ is continuous-valued, 
TPE constructs a probability 
distribution by placing a truncated Gaussian distribution centered at each of
$\vect{x}^k_{k,1:t}$ where $y_t < y^{\ast}$, with standard deviation set to the
greater of the distances to the left and right neighbors. 
\end{itemize}
In the simplest version, each node is independent, so
we can compute $p_t^{<}(\vect{x})$ 
by multiplying individual probabilities at every node.
In the tree-structured version, we only multiply probabilities
along the relevant path, excluding some nodes. 

Another common approach to the surrogate 
is the Gaussian Process \cite{rasmussen,portfolio,bayesiangp}.
Like \newcite{bergstra}, our preliminary experiments found the TPE to
perform favorably.  Further TPE's tree-structured configuration space
is advantageous, because
it allows nested definitions of hyperparameters, which we 
exploit in our experiments (e.g.,
only allows bigrams to be chosen if unigrams are also chosen).

\subsection{Implementation Details}
Because research on SMBO is active, many implementations are publicly
available; we use the HPOlib library \cite{hpolib}.\footnote{\url{http://www.automl.org/hpolib.html}}
The libray takes as input a function $L$, which is treated as a black
box---in our case, a logistic regression trainer that wraps the
LIBLINEAR library \cite{liblinear}, based on the trust region Newton
method \cite{trustnewton}---and a specification of hyperparameters.

\section{Experiments}

Our experiments consider representational choices and hyperparameters
for several text categorization problems.

\subsection{Setup}
We fix our learner $L$ to logistic regression.
We optimize text representation
based on the types of $n$-grams used, the type of weighting scheme,
and the removal of stopwords.
For $n$-grams, we have two parameters, minimum and maximum lengths
($n_{\mathit{min}}$ and $n_{\mathit{max}}$).
(All $n$-gram lengths between the minimum and maximum, inclusive,
are used.)
For weighting scheme, we consider term frequency, tf-idf, and binary schemes.
Last, we also choose whether we should remove stopwords before constructing feature
vectors for each document. 

Furthermore, the choice of representation interacts with the regularizer 
and the training convergence criterion (e.g., more $n$-grams means slower training time).
We consider two regularizers, $\ell_1$ penalty \cite{lasso} or squared $\ell_2$ penalty \cite{ridge}. 
We also have hyperparameters for regularization strength and training convergence tolerance.
See Table~\ref{tbl:hyperparam} for a complete list of hyperparameters
in our experiments. 

Note that even with this limited number of options,
the number of possible combinations is huge (it is actually 
infinite since the regularization strength and
convergence tolerance are continuous values, 
although we can also use sets of possible values), so exhaustive search
is computationally expensive.
In all our experiments for all datasets, we limit ourselves to 30
trials per dataset.  The only preprocessing we applied was downcasing 
(see \S{\ref{sec:discuss}} for discussion about this). 

We always use a development set to evaluate $f(\vect{x})$ during learning
and report the final result on an unseen test set.

\begin{table}[t]
\begin{center}
\begin{tabular}{|l|l|}
\hline \bf Hyperparameter & \bf Values \\ \hline
$n_{\mathit{min}}$ & $\{1,2,3\}$\\
$n_{\mathit{max}}$ & $\{n_{\mathit{min}}, \ldots, 3\}$ \\
weighting scheme & \{tf, tf-idf, binary\} \\
remove stop words? & \{True, False\}\\
\hline
regularization & $\{\ell_1, \ell_2\}$ \\
regularization strength & $[10^{-5}, 10^5]$\\
convergence tolerance & $[10^{-5}, 10^{-3}]$\\
\hline
\end{tabular}
\end{center}
\caption{\label{tbl:hyperparam} The set of hyperparameters considered in our experiments.
The top  half are hyperparameters related to text representation, while the bottom half
are logistic regression hyperparameters, which also interact with the
chosen representation.} 
\vspace{0.1cm}
\end{table}

\subsection{Datasets}
We evaluate our method on five text categorization tasks.
\begin{itemize}
\item Stanford sentiment treebank \cite{socher}: a \emph{sentence-level} sentiment analysis dataset for
movie reviews from the \url{rottentomatoes.com} website. We use the binary classification
task where the goal is to predict whether a review is positive or negative (no neutral reviews).
We obtained this dataset from \url{http://nlp.stanford.edu/sentiment}.
\item Electronics product reviews from Amazon \cite{largeamazon}: this dataset consists of
electronic product reviews, which is a subset of a large Amazon review dataset.
Following the setup of \newcite{riejohnson}, we only use the text section and ignore the summary section.
We also only consider positive and negative reviews.
We obtained this dataset from \url{http://riejohnson.com/cnn_data.html}.
\item IMDB movie reviews \cite{mass}: a binary sentiment analysis dataset of highly polar IMDB movie reviews,
obtained from \url{http://ai.stanford.edu/~amaas//data/sentiment}.
\item Congressional vote \cite{thomas}: transcripts from the U.S. Congressional floor debates.
The dataset only includes debates for controversial bills (the losing side has at least
20\% of the speeches).
Similar to previous work \cite{thomas,ainur}, we consider the task to predict the vote (``yea'' or ``nay'')
for the speaker of each speech segment (speaker-based speech-segment classification).
We obtained it from \url{http://www.cs.cornell.edu/~ainur/sle-data.html}. 
\item 20 Newsgroups \cite{20n}: the 20 Newsgroups dataset is a
benchmark topic classification dataset, we use the publicly available copy at
\url{http://qwone.com/~jason/20Newsgroups}.
There are 20 topics in this dataset.
We derived four topic classification tasks from this dataset.
The first task is to classify documents across all 20 topics.
The second task is to classify related science documents into 
four science topics (\texttt{sci.crypt}, \texttt{sci.electronics},
\texttt{sci.med}, \texttt{sci.med}). \footnote{We were not able to find previous results that are comparable to ours
on the second task; we include them to enable further comparisons in
the future.}
The third and fourth tasks are \texttt{talk.religion.misc} vs.~\texttt{alt.atheism}
and \texttt{comp.graphics} vs.~\texttt{comp.windows.x}.
To consider a more realistic setting, 
we removed header information from each article since they often contain label information. 
\end{itemize}

\begin{table}[t]
\begin{center}\small
\begin{tabular}{|l|r|r|r|}
\hline \bf Dataset & \bf Training & \bf Dev. & \bf Test \\ \hline
Stanford sentiment & 6,920 & 872& 1,821\\
Amazon electronics & 20,000  & 5,000 & 25,000\\
IMDB reviews & 20,000 & 5,000 & 25,000 \\
Congress vote & 1,175 & 113 & 411\\
20N all topics & 9,052& 2,262& 7,532\\
20N all science & 1,899& 474& 1,579\\
20N atheist.religion & 686 & 171 & 570\\
20N x.graphics & 942& 235 & 784\\
\hline
\end{tabular}
\end{center}
\caption{\label{tbl:datasets} Document counts.}
\end{table}

These are standard datasets for evaluating text categorization models,
where benchmark results are available.
In total, we have eight tasks, of which four are
sentiment analysis tasks  and four  are
topic classification tasks.
See Table~\ref{tbl:datasets} for descriptive statistics of our datasets.

\begin{table*}[t]
\begin{center}\small
\begin{tabular}{|l|r||r|r|r|r||r|r|r|r|}
\hline \bf Dataset & \bf Acc. & \bf $n_{\mathit{min}}$ & \bf $n_{\mathit{max}}$ & Weighting & Stop. & Reg. & Strength & Conv. \\ \hline
Stanford sentiment & 82.43 &  1 & 2 & tf-idf & F& $\ell_2$& 10 & 0.098\\
Amazon electronics & 91.56 & 1 & 3 & binary& F& $\ell_2$& 120& 0.022\\
IMDB reviews & 90.85 & 1 & 2& binary& F& $\ell_2$& 147& 0.019\\
Congress vote & 78.59 & 2 & 2 & binary & F & $\ell_2$ & 121& 0.012\\
20N all topics & 87.84 & 1&2 & binary &F & $\ell_2$ & 16& 0.008\\
20N all science & 95.82& 1&2 & binary &F & $\ell_2$ & 142& 0.007\\
20N atheist.religion & 86.32 &1 &2 & binary & T & $\ell_1$ & 41& 0.011\\
20N x.graphics & 92.09 & 1 & 1 & binary & T & $\ell_2$ & 91& 0.014\\
\hline
\end{tabular}
\end{center}
\caption{\label{tbl:results} Classification accuracies and the best hyperparameters for each of the dataset in our experiments.
``Acc'' shows accuracies for our logistic regression model.
``Min'' and ``Max'' correspond to the min $n$-grams and max $n$-grams respectively.
``Stop.'' is whether we perform stopwords removal or not.
``Reg.'' is the regularization type, 
``Strength'' is the regularization strength, 
and ``Conv.'' is the convergence tolerance. 
For regularization strength, we round it to the nearest integer for readability.
}
\end{table*}

\subsection{Baselines}
For each dataset, we select supervised, non-ensemble classification methods from
previous literature as baselines.  
In each case, we emphasize comparisons with the
best-published linear method (often an SVM with a linear kernel with representation
selected by experts) and the best-published
method overall.
In the followings, ``SVM'' always means ``linear SVM''.
All methods were trained and evaluated on the same
training/testing data splits; in cases where standard development sets
were not available, we used a random 20\% of the
training data as a development set.

\subsection{Results}
We summarize the hyperparameters selected by our method, and the
accuracies achieved (on test data) 
in Table~\ref{tbl:results}.  We discuss comparisons to baselines for each dataset in
turn.

\paragraph{Stanford sentiment treebank (Table~\ref{tbl:sst}).}
Our logistic regression model outperforms the baseline SVM reported by \newcite{socher},
who used only unigrams but
did not specify the weighting scheme for their SVM baseline.
While our result is still below the state-of-the-art based on the
the recursive neural tensor networks \cite{socher} and the
paragraph vector \cite{paragraphvector},
we show that logistic regression
is comparable with recursive and matrix-vector neural networks \cite{socherrnn,sochermvrnn}.

\begin{table}[h]
\begin{center}
\begin{tabular}{|l|r|}
\hline \bf Method & \bf Acc. \\ \hline
Na\"{i}ve Bayes & 81.8 \\
SVM & 79.4 \\
Vector average & 80.1 \\
Recursive neural networks & 82.4\\
\hline
\textbf{LR (this work)} & 82.4 \\ \hline
Matrix-vector RNN & 82.9\\
Recursive neural tensor networks & 85.4\\
Paragraph vector & 87.8 \\

\hline
\end{tabular}
\end{center}
\caption{\label{tbl:sst} Comparisons on the Stanford sentiment treebank dataset.
Scores are as reported by \newcite{socher} and \newcite{paragraphvector}.
}
\end{table}

\paragraph{Amazon electronics (Table~\ref{tbl:elec}).}
The best-performing methods on this dataset are based on convolutional
neural networks \cite{riejohnson}.\footnote{These are fully connected
  neural networks with a rectifier activation function, trained under
  $\ell_2$ regularization with stochastic gradient descent.}  Our method is on par with the
second-best of these, outperforming all of the reported feed-forward
neural networks and SVM variants Johnson and Zhang used as baselines.
They varied the representations, and used log term frequency
and normalization to unit vectors as the weighting scheme, after
finding that this outperformed term frequency.  Our method achieved
the best performance with binary weighting, which they did not
consider. 

\begin{table}[ht]
\begin{center}
\begin{tabular}{|l|r|}
\hline \bf Method & \bf Acc. \\ \hline
SVM-unigrams & 88.62\\
SVM-$\{1,2\}$-grams & 90.70\\
SVM-$\{1,2,3\}$-grams & 90.68\\
NN-unigrams & 88.94 \\
NN-$\{1,2\}$-grams & 91.10 \\
NN-$\{1,2,3\}$-grams & 91.24 \\
\hline
\bf LR (this work) & 91.56 \\
\hline 
Bag of words CNN & 91.58 \\
Sequential CNN & 92.22 \\
\hline
\end{tabular}
\end{center}
\caption{\label{tbl:elec} Comparisons on the Amazon electronics dataset.
Scores are as reported by \newcite{riejohnson}.}
\end{table}

\paragraph{IMDB reviews (Table~\ref{tbl:imdb}).}
The results parallel those for Amazon electronics; our method comes
close to convolutional neural networks \cite{riejohnson}, which are
state-of-the-art.\footnote{As noted, semi-supervised and ensemble methods are
  excluded for a fair comparison.}
It outperforms SVMs and feed-forward neural networks, the
restricted Boltzmann machine approach presented by \newcite{dahl}, 
and compressive feature learning \cite{compressive}.\footnote{This
  approach
is based on minimum description length, using unlabeled data to select
a set of higher-order $n$-grams to use as features.  It is technically
a semi-supervised method.  The results we compare to use logistic
regression with elastic net regularization and heuristic normalizations.}

\begin{table}[h]
\begin{center}
\begin{tabular}{|l|r|}
\hline \bf Method & \bf Acc. \\ \hline
SVM-unigrams & 88.69 \\
SVM-$\{1,2\}$-grams & 89.83 \\
SVM-$\{1,2,3\}$-grams & 89.62 \\
RBM & 89.23 \\
NN-unigrams & 88.95 \\
NN-$\{1,2\}$-grams & 90.08 \\
NN-$\{1,2,3\}$-grams & 90.31 \\
Compressive feature learning & 90.40 \\
LR-$\{1,2,3,4,5\}$-grams & 90.60 \\ 
\hline
\bf LR (this work) & 90.85 \\ \hline
Bag of words CNN & 91.03 \\
Sequential CNN & 91.26 \\
\hline
\end{tabular}
\end{center}
\caption{\label{tbl:imdb} Comparisons on the IMDB reviews dataset.
SVM results are from \newcite{wangmanning}, the RBM (restricted
Bolzmann machine) result is from \newcite{dahl},
NN and CNN results are from \newcite{riejohnson},
and LR-$\{1,2,3,4,5\}$-grams and compressive feature learning
results are from \newcite{compressive}.
}
\end{table}

\paragraph{Congressional vote (Table~\ref{tbl:convote}).}
Our method outperforms the best reported results of \newcite{ainur},
which use a multi-level structured model based on a latent-variable
SVM.  We show comparisons to two well-known but weaker baselines, as well.

\begin{table}[h]
\begin{center}
\begin{tabular}{|l|r|r|}
\hline \bf Method & \bf Acc. \\ \hline
SVM-link & 71.28 \\
Min-cut   & 75.00 \\
SVM-SLE & 77.67 \\
\hline
\bf LR (this work) & 78.59 \\
\hline
\end{tabular}
\end{center}
\caption{\label{tbl:convote} Comparisons on the U.S. congressional vote dataset.
SVM-link exploits link structures \cite{thomas}; the min-cut result is from \newcite{bansal};
and SVM-SLE result is reported by \newcite{ainur}.
}
\end{table}

\paragraph{20 Newsgroups:  all topics (Table~\ref{tbl:20n}).}
Our method outperforms state-of-the-art methods including the
distributed structured output model \cite{srikumar}.\footnote{This
  method was designed for structured prediction, but
  \newcite{srikumar} also applied it to classification.  It attempts
  to learn a distributed representation for features and for labels.
  The authors used unigrams and did not elaborate the weighting scheme.}
The strong logistic regression baseline from \newcite{compressive}
uses all 5-grams, heuristic normalization, and elastic net regularization; 
our method found that
unigrams and bigrams, with binary weighting and $\ell_2$ penalty, achieved far better results.

\begin{table}[h]
\begin{center}
\begin{tabular}{|l|r|r|}
\hline \bf Method & \bf Acc. \\ \hline
Discriminative RBM & 76.20 \\
Compressive feature learning & 83.00 \\
LR-$\{1,2,3,4,5\}$-grams & 82.80 \\
Distributed structured output & 84.00 \\
\hline
\bf LR (this work) & 87.84 \\
\hline
\end{tabular}
\end{center}
\caption{\label{tbl:20n} Comparisons on the 20 Newsgroups dataset for
classifying documents into all topics.
The disriminative RBM result is from \newcite{drbm};
compressive feature learning and LR-5-grams results
are from \newcite{compressive},
and the distributed structured output result is from \newcite{srikumar}.
}
\end{table}

\paragraph{20 Newsgroups:  \texttt{talk.religion.misc}
  vs.~\texttt{alt.atheism} and \texttt{comp.graphics} vs.~\texttt{comp.windows.x}}  \newcite{wangmanning} report a
bigram na\"{i}ve Bayes model achieving 85.1\% and 91.2\% on these
tasks, respectively.\footnote{They also report a na\"{i}ve
  Bayes/SVM ensemble achieving 87.9\% and 91.2\%.}
Our method achieves 86.3\% and 92.1\% using slightly different setups
(see Table~\ref{tbl:results}).

\begin{figure*}[t]
\centering
\includegraphics[scale=0.33]{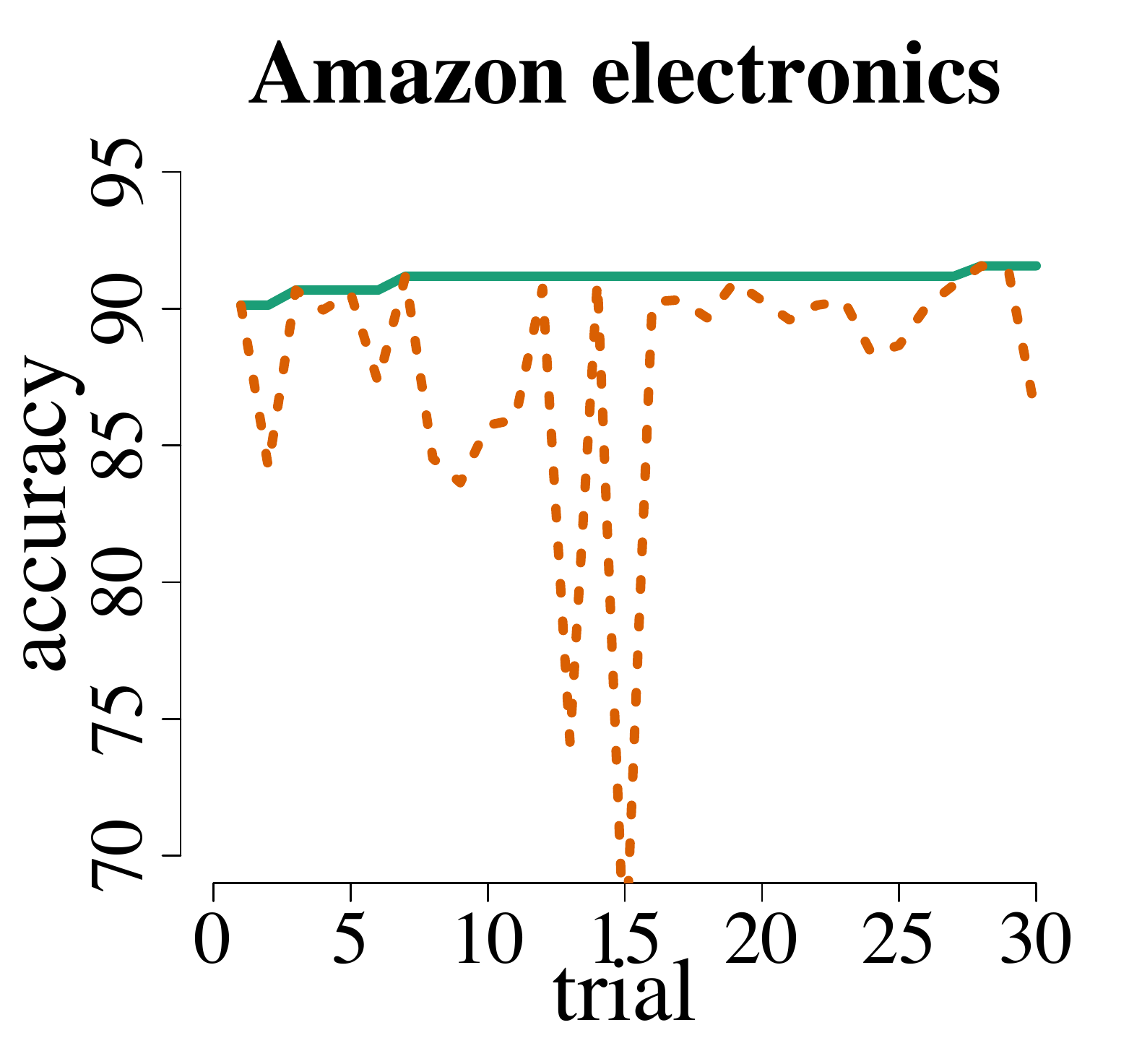}
\includegraphics[scale=0.33]{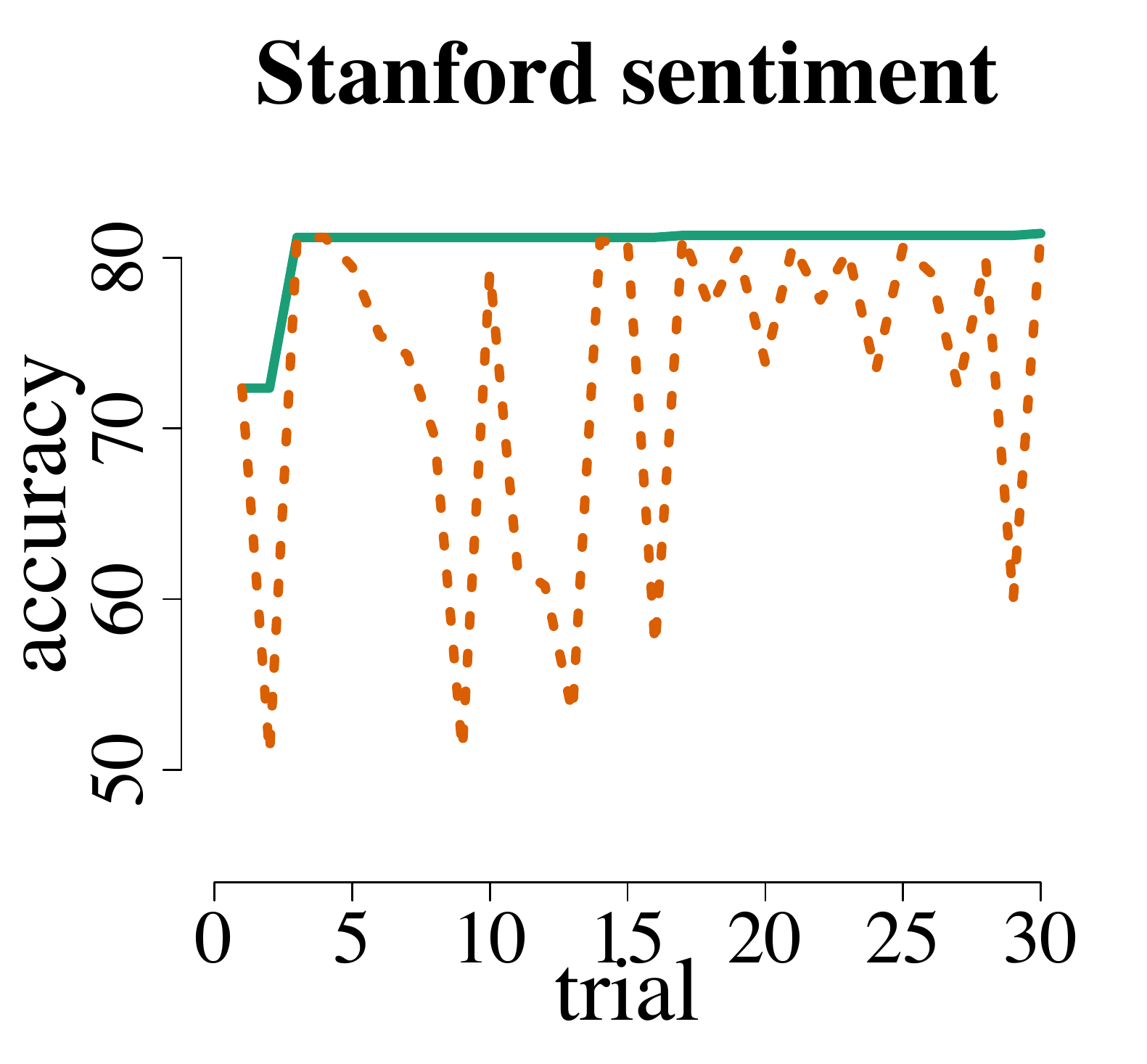}
\includegraphics[scale=0.32]{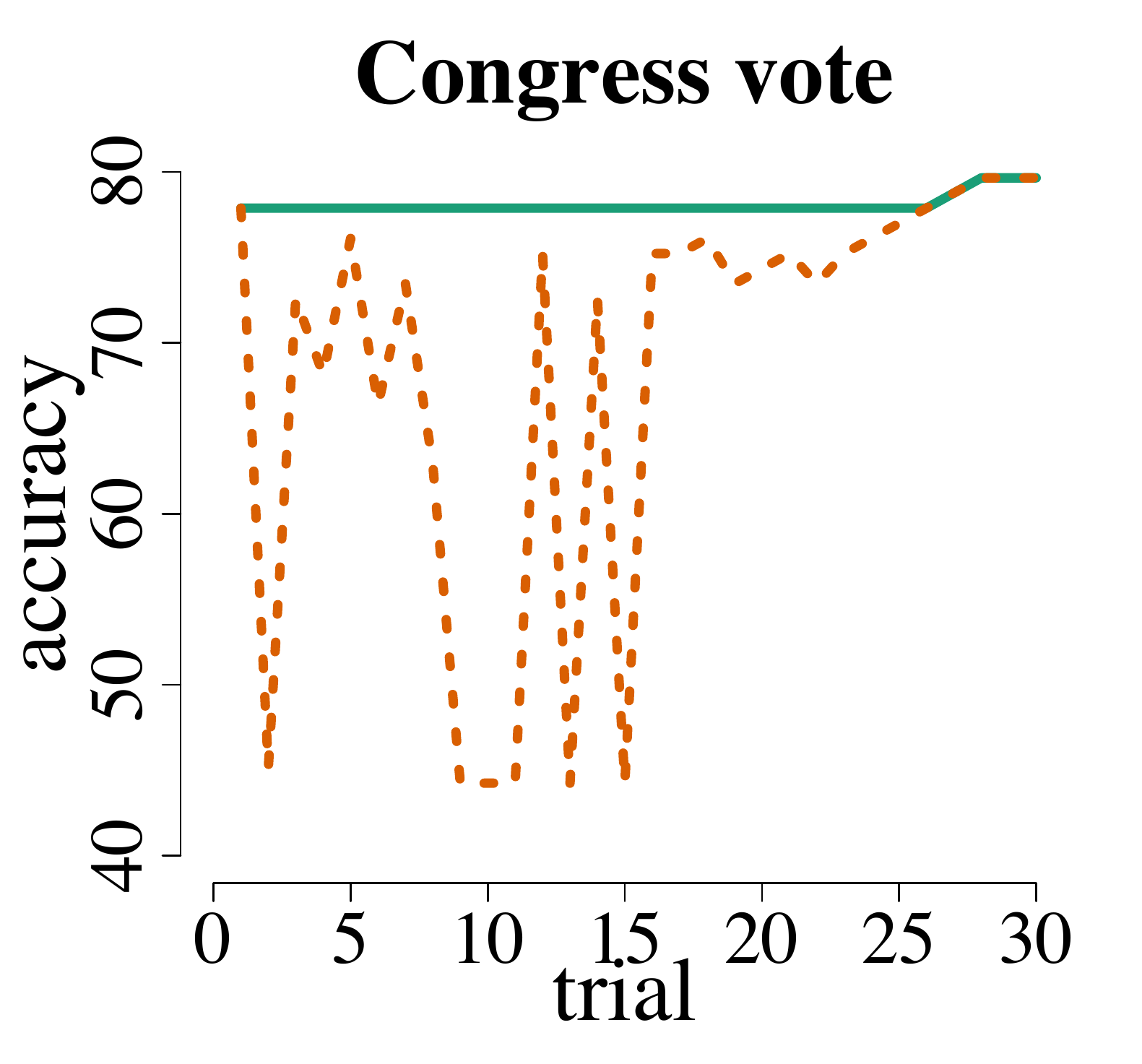}
\caption{
\label{fig:imdb} Classification accuracies on development data for
Amazon electronics (left), Stanford sentiment treebank (center), and congressional vote (right) datasets.
In each plot, the green solid line indicates the best accuracy found so far, while the dotted orange line
shows accuracy at each trial. We can see that in general the model is able to
obtain reasonably good representation in 30 trials.
}
\end{figure*}

\section{Discussion}
\label{sec:discuss}
\paragraph{Raw text as input and other hyperparameters.}
Our results suggest that seemingly mundane representation choices can
raise
the performance of simple linear models to be comparable with much
more sophisticated models.  Achieving these results is not a matter of deep expertise
about the domain or engineering skill; the choices can be automated.
  Our experiments only
considered
logistic regression with downcased text; more choices---stemming,
count thresholding, normalization of numbers, etc.---can be offered to
the optimizer, as can additional feature options like gappy $n$-grams.

As NLP becomes more widely used in applications, we believe that
automating these choices will be very attractive for those who need to
train a high-performance model quickly.

\paragraph{Optimized representations.}
For each task, the chosen representation is different.
Out of all possible hyperparameter choices in our experiments (Table~\ref{tbl:hyperparam}),
each of them is used by at least one of the datsets (Table~\ref{tbl:results}). 
For example, on the Congressional Vote dataset, we only need to use bigrams,
whereas on the Amazon electronics dataset we need to use unigrams, bigrams, and trigrams.
The binary weighting scheme works well for most of the datasets, except the sentence-level
sentence analysis task, where the tf-idf weighting scheme was selected.
$\ell_2$ regularization was best in all cases but one.

We do not believe that an NLP expert would be likely to make these
particular choices, except through the same kind of trial-and-error
process our method automates efficiently.  Often, we believe, researchers in NLP
make initial choices and stick with them through all experiments (as
we have admittedly done with logistic regression). 
Optimizing over more of these choices will give stronger baselines.

\paragraph{Training time.}
We ran 30 trials for each dataset in our experiments.
Figure~\ref{fig:imdb} 
shows each trial accuracy and the best accuracy on development data
as we increase the number of trials for three datasets.
We can see that 30 trials are generally enough 
for the model to obtain good results, although the search space is large. 

In the presence of unlimited computational resources,
Bayesian optimization is slower than grid search on all
hyperparameters, since the latter is easy to parallelize.
This is not realistic in most research and development environments,
and it is certainly impractical in increasingly widespread
instances of
personalized machine learning.
The Bayesian optimization approach that we use in our experiments
is performed sequentially. It
attempts to \emph{predict}
what set of hyperparameters we should try next
based on information from previous trials.
There has been work to parallelize Bayesian optimization,
making it possible to leverage the power of multicore architectures \cite{bayesiangp,parallel1,parallel2}. 

\paragraph{Transfer learning and multitask setting.}
We treat each dataset independently and create a separate model
for each of them. It is also possible to learn from
previous datasets (i.e., transfer learning) or to learn
from all datasets simultaneously (i.e., multitask learning) to 
improve performance.
This has the potential to reduce the number of trials required
even further. See \newcite{bardenet}, \newcite{multitask}, and \newcite{yogatamamann2014}
for how to perform Bayesian optimization in these settings. 

\paragraph{Beyond linear models.}
We use logistic regression as our classification model,
and our experiments show how simple linear models can be
competitive with more sophisticated models given the right
representation.  Other models, can be considered, of course, as can ensembles \cite{yogatamamann2014}.
Increasing the number of options may lead to a need for more trials, and
evaluating $f(\vect{x})$ (e.g., 
training the neural network) will take longer for more sophisticated models.
We have demonstrated, using one of the simplest classification models
(logistic regression), that even simple choices about text representation can matter quite a lot.

\paragraph{Structured prediction problems}
Our framework could also be applied to structured prediction
problems. 
For example, in part-of-speech tagging, the set of features
can include character $n$-grams, word shape features, and word type features.
The optimal choice for different languages is not always the same,
our approach can automate this process.

\paragraph{Beyond supervised learning.}
Our framework could also be extended to unsupervised and
semi-supervised models.
For example, in document clustering (e.g., $k$-means), 
we also need to construct representations for documents.
Log-likelihood might serve as a performance function.  A range of
random initializations might be considered. 
Investigation of this approach 
for nonconvex problems like clustering is an exciting area for future
work.

\section{Conclusion}
We used a Bayesian optimization approach to optimize choices about
text representations for various categorization problems.
Our sequential model-based optimization technique
identifies settings for a standard linear model (logistic regression)
that are competitive with far more sophisticated state-of-the-art
methods on topic classification and sentiment analysis.
Every task and dataset has its own optimal choices; though relatively
uninteresting to researchers and not directly linked to domain or
linguistic expertise, these choices have a big effect on
performance.
We see our approach as a first step towards black-box NLP systems that
work with raw text and do not require manual tuning.

\section*{Acknowledgements}
This work was supported by the Defense Advanced Research Projects Agency through grant FA87501420244 
and computing resources provided by Amazon. 

\bibliographystyle{acl}
\bibliography{acl2015}

\end{document}